\documentclass[runningheads]{llncs}
\newcommand{\myheader}{Preprint, published at the \textit{International Conference on Artificial Neural Networks (ICANN)} 2020}

\usepackage{atbegshi,picture}
\usepackage{datetime}
\usepackage{lastpage}

\newcommand{\myleftstd}{1.5cm}
\usepackage{url}

\ifthenelse{\isodd{\thepage}}
 {\newcommand{\myleftmargin}{\oddsidemargin+\myleftstd}}
 {\newcommand{\myleftmargin}{\evensidemargin+\myleftstd}}

\AtBeginShipoutNext {
 \AtBeginShipoutUpperLeft{%
  \put(\dimexpr 0.0\myleftmargin\relax,-1.2cm){\parbox{\paperwidth}{\footnotesize\sffamily\noindent{\centering \myheader \hfill}}}%
}}
\usepackage[numbers,sectionbib]{natbib}
\usepackage[capitalise]{cleveref}
\usepackage{subcaption}
\usepackage{graphicx}

\begin{document}
\title{Curious Hierarchical Actor-Critic Reinforcement Learning}

\author{Frank R{\"o}der$^*$ \and
Manfred Eppe$^*$ \and
Phuong D.H. Nguyen \and
Stefan Wermter}

\authorrunning{F. R{\"o}der et al.}

\institute{Department of Informatics \newline Knowledge Technology Institute
  \newline Universit{\"a}t Hamburg, Hamburg, Germany
  \email{\{3roeder,eppe,pnguyen,wermter\}@informatik.uni-hamburg.de}}

\maketitle

\begin{abstract}
Hierarchical abstraction and curiosity-driven exploration are two common paradigms in current reinforcement learning approaches to break down difficult problems into a sequence of simpler ones and to overcome reward sparsity.
However, there is a lack of approaches that combine these paradigms, and it is currently unknown whether curiosity also helps to perform the hierarchical abstraction.
As a novelty and scientific contribution, we tackle this issue and develop a method that combines hierarchical reinforcement learning with curiosity.
Herein, we extend a contemporary hierarchical actor-critic approach with a forward model to develop a hierarchical notion of curiosity. We demonstrate in several continuous-space environments that curiosity can more than double the learning performance and success rates for most of the investigated benchmarking problems. We also provide our source code~\footnote{\url{https://github.com/knowledgetechnologyuhh/goal_conditioned_RL_baselines}} and a supplementary video~\footnote{\url{https://www2.informatik.uni-hamburg.de/wtm/videos/chac_icann_roeder_2020.mp4}}.
\end{abstract}
$^*$ Equal contribution
\section{Introduction}\label{sec:introduction}

\begin{figure}[ht]
    \centering
    \includegraphics[width=\linewidth]{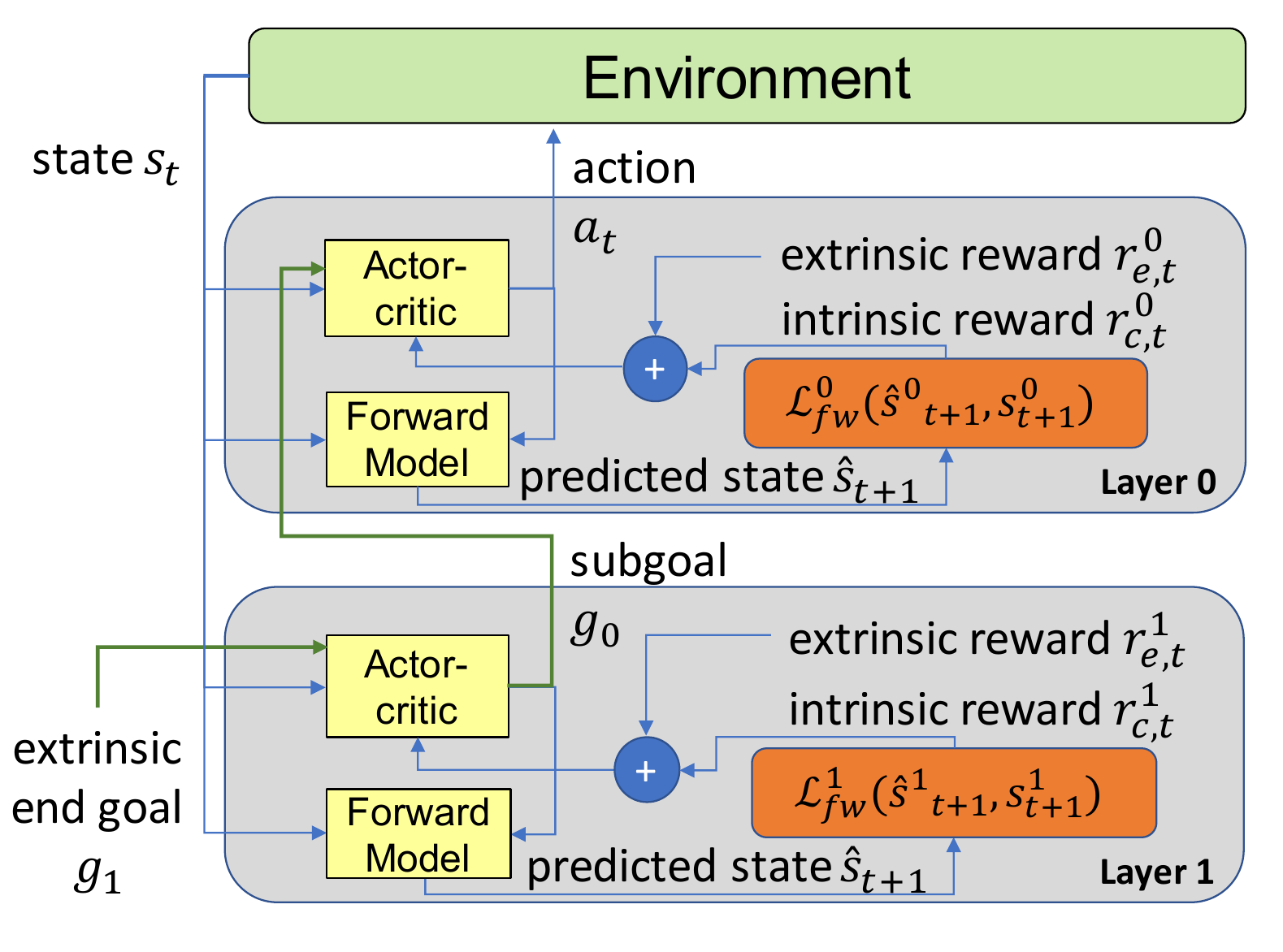}
    \caption{The CHAC Architecture with two layers of hierarchy. A forward model is employed to compute the prediction error $\mathcal{L}^i_{fw}(\hat{s}^i_{t+1}, s^i_{t+1})$, which provides an additional curiosity-based reward $r^i_{c,t}$ for the layer $i$ of hierarchy. This intrinsic reward is added to the extrinsic reward $r^i_{e,t}$ to train the actor-critic.}
    \label{fig:chac_architecture}
    \vspace{-0.8cm}
\end{figure}

A general problem for reinforcement learning (RL) is sparse rewards.
For example, tasks as simple as drinking water involve a complex sequence of motor commands, and only upon completion of this complex sequence, a reward is provided, which destabilizes the learning of value functions.
Hierarchical reinforcement learning (HRL) partially alleviates this issue by decomposing difficult tasks into simpler subtasks, providing additional intrinsic rewards upon completion of the subtasks.
Therefore, HRL is a major step towards human-like cognition~\cite{Pezzulo2018_HierarchicalActiveInference} and decision-making~\cite{Botvinick2014}.
There exists a considerable body of research demonstrating that hierarchical architectures provide a significant performance gain compared to non-hierarchical architectures by performing such abstractions \cite{Eppe2019_planning_rl,Levy2019_Hierarchical,Vezhnevets2017}.

However, HRL does not completely eliminate the problem of reward sparsity. By adding intrinsic rewards for achieving subtasks, it rather transforms the problem of reward sparsity into the problem of selecting the appropriate subgoals or subtasks. Learning the subgoal or subtask-selection still suffers from reward sparsity. So how can we improve the learning of subtask selection under sparse rewards?

Current RL literature offers two commonly used methods for overcoming rewards sparsity that we will investigate to address this question. The first method is hindsight experience replay (HER) \cite{Andrychowicz2017}. The idea behind HER is to pretend in hindsight that the final state of a rollout was the goal of the rollout, regardless of whether it was actually the original one. This way, unsuccessful rollouts get rewarded by considering in hindsight that they were successful.
In recent work, \citet{Levy2019_Hierarchical} have successfully combined HER with a hierarchical actor-critic reinforcement learning approach, demonstrating a significant performance gain for several continuous-space environments.
The second method to densify rewards is curiosity. Existing curiosity-based approaches in non-hierarchical reinforcement learning  (e.g.~\cite{Hafez2017,Pathak2017_forward_model_intrinsic}) provide additional rewards when the agent is surprised.
Following research around \citet{Friston2011}, the notion of surprise is based on the prediction error of an agent's internal forward model.
That is, the agent is surprised when its internal prediction of the world dynamics does not coincide with its actual dynamics.

There exists a significant amount of recent approaches on hierarchical reinforcement learning (e.g.
\cite{Bacon2017_OptionCritic,Jaderberg2017_unreal,Jiang2019_HRL_Language_Abstraction,Kulkarni2016,Levy2019_Hierarchical,Nachum2018_HIRO,Vezhnevets2017}). We are also aware of significant recent improvements in curiosity-driven non-hierarchical reinforcement learning (e.g. \cite{Alet2020Meta-learningAlgorithms,Burda2019_ICLR,Burda2018,Colas2019,Forestier2016,Hafez2017,Hester2017,Pathak2017_forward_model_intrinsic,Watters2019_COBRA}).
However, despite significant evidence from Cognitive Sciences, suggesting that curiosity is a hierarchical phenomenon \cite{Pezzulo2018_HierarchicalActiveInference}, there exist no functional computational models to verify this hypothesis.

In this paper, we address this lack and ask the following \textbf{central research question}: \emph{To what extent can we alleviate reward-sparsity and improve the learning performance of hierarchical actor-critic reinforcement learning with a hierarchical curiosity mechanism?}

We address this question by extending the hierarchical actor-critic approach by \citet{Levy2019_Hierarchical} with a reward signal that fosters the agent's curiosity. We extend the approach with \citeauthor{Friston2011}'s proposal to model surprise based on prediction errors \cite{Friston2011} and provide the agent with intrinsic rewards if it is surprised (cf. \Cref{fig:chac_architecture}).
 As a novelty and scientific contribution, we are the first to present a computational model that combines curiosity with hierarchical reinforcement learning, and that considers also hindsight experience replay as an additional method to overcome reward sparsity. We refer to our method as Curious Hierarchical Actor-Critic (CHAC) and evaluate our approach in several continuous-space benchmark environments.

\section{Background and Related Work}\label{sec:basics}
Our research integrates hierarchical reinforcement learning with a curiosity and surprise mechanism inspired by the principle of active inference \cite{Friston2011}.
In the following, we provide the background of these mechanisms and methods.

\subsection{Reinforcement Learning}
\label{sec:rl}
Reinforcement learning (RL) involves a Markov Decision Process (MDP) to maximize
the long-term expected reward. An MDP is defined as a tuple, $\langle
\mathcal{S},\mathcal{A},\mathcal{R},\mathcal{T},\gamma \rangle$, where
$\mathcal{S}$ is a set of states, $\mathcal{A}$ is a set of actions,
$\mathcal{R}: \mathcal{S} \times \mathcal{A}$ is a reward function,
$\mathcal{T}:\mathcal{S} \times \mathcal{A} \mapsto
Pr(\mathcal{S})=p(s_{t+1}|s_t,a_t)$ is a transition probability of reaching
state $s_{t+1}$ from the current state $s_t$ when executing action $a_t$,
and $\gamma \in [0,1)$ is a discount factor, indicating how much the agent
prefers short-term to long-term rewards.
In our setting, the agent takes actions drawn from a probability distribution over action, a policy, denoted $\pi(a|s): \mathcal{S} \mapsto \mathcal{A}$.
The goal of the agent is to take actions that maximize long-term expected reward.
In this work, we employ the Deep Deterministic Policy Gradient (DDPG) algorithm~\cite{Lillicrap2016_DDPG} for the policy learning. DDPG is a model-free off-policy actor-critic algorithm, which combines the Deterministic Policy Gradient (DPG) algorithm~\cite{Silver2014DeterministicAlgorithms} with Deep Q-network (DQN)~\cite{Mnih2015}. This enables agent with DDPG to work in continuous space while learning with large, non-linear function approximators more stably and efficiently.
In \Cref{sec:model}, we define how this non-hierarchical notion of reinforcement learning is extended to the hierarchical actor-critic case.

\subsection{Curiosity-Driven Exploration}\label{sub:curiosity}
\citet{Friston2011} describe surprise as ``the improbability of sampling some signals, under a generative model of how those signals were caused.''. Hence, curiosity can be achieved by maximizing surprise, i.e., by maximizing the probability of sampling signals that do not coincide with the predictions by the generative model \cite{Butz2016,Friston2011}\footnote{Note that curiosity is a broad term and there exist other rich notions of curiosity \cite{Gottlieb2018_curiority_nature}. However, for this paper we focus on the well-defined and established notion of curiosity as maximizing a function over prediction errors.}.

A common method realizing this in practical reinforcement learning applications is to define a generative forward model $f_{fw} : \mathcal{S} \times \mathcal{A} \mapsto \mathcal{S}$ that maps states and actions to successive states.
One can then use the forward model to implement surprise as a function of the error between the successive states predicted by the model and the actual successive states.
This strategy and derivatives thereof have been successfully employed in several non-hierarchical reinforcement learning approaches \cite{Alet2020Meta-learningAlgorithms,Burda2019_ICLR,Burda2018,Butz2016,Forestier2016,Hafez2017,Hester2017,Pathak2017_forward_model_intrinsic,Schillaci2016,Schmidhuber2010,Watters2019_COBRA}.

For example, \citet{Pathak2017_forward_model_intrinsic} propose an Intrinsic Curiosity Module, introducing an additional internal reward that is defined as the squared error of the predictions generated by a forward model. Similarly, \citet{Hafez2017} implement surprise as the absolute error of a set of forward models, and \citet{Watters2019_COBRA} use the squared error as a reward signal.

\section{Curious Hierarchical Actor-Critic}\label{sec:model}
The hierarchical actor-critic (HAC) approach by \citet{Levy2019_Hierarchical} has shown great potential in continuous-space environments. At the same time, there exists extensive research \cite{Hafez2017,Pathak2017_forward_model_intrinsic} showing how curious agents striving to maximize their surprise can improve their learning performance. In the following, we describe how we combine both paradigms.

\subsection{Hierarchical Actor-Critic}
\label{sec:model:hac}
Hierarchical actor-critic (HAC) \cite{Levy2019_Hierarchical} is a framework that enables agents to learn a nested hierarchy of policies. It uses hindsight experience replay (HER) \cite{Andrychowicz2017} to alleviate reward-sparsity.
Each layer of the hierarchy learns to solve a subproblem defined by the spaces and a transition function of the layers below: It produces actions that are subgoals for the next lower level.
The highest layer receives the current state and the overall extrinsic goal as input.
The lowest layer produces motor commands that are executable by the agent in the environment. HAC involves the following three kinds of state transitions that implement HER in a hierarchical setting.

\emph{Hindsight Goal Transitions} are akin to the transitions in the non-hierarchical HER method: After a rollout has completed, the agent pretends in hindsight that the actually achieved state was the goal state.
They enable the critic function to encounter at least one sparse reward after a sequence of actions.
\emph{Hindsight Action Transitions}:
These additional state transitions
are generated by pretending in hindsight that the action provided as subgoal to the low-level layer has been achieved.
This alleviates the slow learning of a hierarchical layer due to the sparsity in achieving the subgoal provided by a higher level.
As a result, HAC can learn multiple levels of policies in parallel, even if the lower-level policies are not yet fully trained.
\emph{Subgoal Testing Transitions}
foster the generation of subgoals that are actually achievable by the low-level layer. They are used to test whether subgoals can be achieved and penalize a subgoal that could not be reached.
Since difficult subgoals are penalized in the beginning of the training, but not anymore when the agent's performance has improved, subgoal testing mechanism provides HAC with a method to automatically generate a curriculum.

We build our approach on these transitions using the following formal framework: We define a hierarchy of $k$ layers with each containing an actor-critic network and a replay buffer to store experiences. Here the RL setting (cf.~\Cref{sec:rl}) is expanded for hierarchical agents.
Each layer $\Pi_i$ of the hierarchy is described as a Universal Markov Decision Process (UMDP), an extension of MDP with an additional set of goals
by applying universal value function approximator (UVFA)~\cite{Schaul2015}. An UMDP is a tuple $\mathcal{U}_i=\langle \mathcal{S}_i,\mathcal{G}_i,\mathcal{A}_i,\mathcal{T}_i,\mathcal{R}_i,\gamma_i \rangle$ containing the state space $\mathcal{S}_i$, the goal space $\mathcal{G}_i$, the action space $\mathcal{A}_i$, the transition probability function $\mathcal{T}_i=p_i(s^i_{t+1}|a^i,s^i_{t})$, the reward function $\mathcal{R}_i$, and the discount rate $\gamma_i \in [0, 1)$ for each layer $i$.
The state space of each layer is identical to the original, namely $\mathcal{S}_i = \mathcal{S}$.
The produced subgoals by the policy $\pi_i: \mathcal{S} \times \mathcal{G}_i \mapsto \mathcal{A}_i$ of each layer are within $\mathcal{S}$, and therefore $\mathcal{G}_i = \mathcal{S}$. The action space is equal to the goal space of the next lower layer, except the lowest one, thus $\mathcal{A}_i = \mathcal{S}, \; i > 0$. Only in the lowest layer, we execute the so-called primitive actions of the agent within the environment and therefore have $\mathcal{A}_0 = \mathcal{A}$~\cite{Levy2019_Hierarchical}.

\subsection{Combining Hierarchical Actor-Critic with Curiosity}
To combine HAC with curiosity-based rewards, we implement a forward model based on a multi-layered perceptron that learns to predict the successive state $\hat{s}_{t+1}$ given the current state $s_t$ and an action $a_t$ at time $t$.
Formally, this mapping is given as follows, with the model parameters $\theta$:
\begin{equation}
\label{eq:forward}
  f_{fw}(s_t, a_t; \theta) \Rightarrow \hat{s}_{t+1}
\end{equation}
An action $a^i_t$ produced by a policy $\pi_i$ of the layer $i$ (except the bottom layer, where $i=0$) at time $t$ is a subgoal for the subsequent level.
We implement one forward model $f^i_{fw}(s_t,a^i_t;\theta^i)$ per layer. That is, we define a forward model not only for the primitive action $a^{i=0} \in \mathcal{A}$ in the lowest layer but also for the subgoal action $a^i \in \mathcal{A}_i=\mathcal{S}$ in the higher layers.
The learning objective for training the forward model is to minimize the prediction loss, defined as:
\begin{equation}
  \mathcal{L}^i_{fw}(\hat{s}^i_{t+1}, s^i_{t+1}) = \frac{(s^i_{t+1} - \hat{s}^i_{t+1})^2}{2}.
\end{equation}
Similar to the approach by~\citet{Pathak2017_forward_model_intrinsic}, the forward model's error of the layer $i$ is used to realize the curiosity-based bonus, denoted as $r^i_{c,t}$.
We calculate the mean-squared-error as follows:

\begin{equation}
\label{eq:r_c}
  r^i_{c,t} = \frac{(s^i_{t+1} - \hat{s}^i_{t+1})^2}{2}
\end{equation}

The regular extrinsic rewards (from the environment) are defined in the range of $[-1,0]$, hence we need to normalize the curiosity reward $r^{i}_{t,c}$ resulted of \Cref{eq:r_c}. The normalization of the curiosity reward is conducted with respect to the maximum and minimum values of the curiosity level in the whole history (stored in a buffer), $r^i_{c,max}$ and $r^i_{c,min}$ respectively, as follows:

\begin{equation}
    r^i_{c,t} = \frac{r^i_{c,t} - r^i_{c,min}}{r^i_{c,max} - r^i_{c,min}} - 1
    \label{eq:normalize_ri}
\end{equation}
In other words, if the prediction error is high, corresponding to high curiosity, the normalized value will be close to~$0$, otherwise, it is close to~$-1$.

The total reward $r^i_t$ at time $t$ that layer $i$ receive, given the extrinsic reward $r^i_{e,t}$ and the curiosity reward $r^i_{c,t}$, is controlled by the hyper-parameter $\eta$ as follows:
\begin{equation}
\label{eq:reward}
  r^i_{t} = \eta \cdot r^i_{e,t}  + (1 - \eta) \cdot r^i_{c,t}
\end{equation}

This part is crucial in determining the balance of
changing the reward, since $r^i_t = r^i_{e,t}$ if $\eta = 1$, which is identical to HAC. We further elaborate on the different values of
$\eta$ in \Cref{sec:analysis}.

\subsection{Architecture and Training}\label{sub:arch_train}

\begin{figure}[ht]
    \centering
    \includegraphics[width=\linewidth]{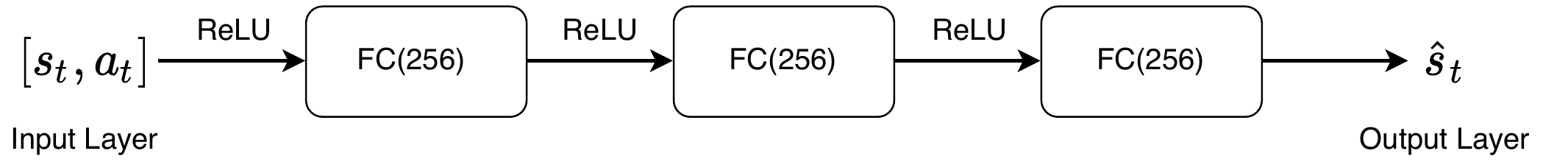}
    \caption{Forward Model Architecture}
    \label{fig:chac_fw_model_architecture}
\end{figure}

We implement the forward model (of each hierarchical layer $i$) as a multilayer perceptron (MLP), receiving the concatenated current state $s_t$ and action $a_t$, to generate a prediction for the successor state $\hat{s}_{t+1}$ as output (cf. \Cref{eq:forward}).
For all experiments in this paper (see \Cref{sec:analysis}), we use an MLP with 3 hidden layers of size 256 (cf. \Cref{fig:chac_fw_model_architecture}) to learn the forward model from the agent's experiences.
Experimentally, we found that this setting yields the best performance results.
Following~\citet{Levy2019_Hierarchical}, we also realize the actor and critic networks with MLPs of 3 hidden layers of size $64$.

Both the forward model and actor-critic are trained consecutively with a learning rate of $0.001$ using the ADAM optimizer~\cite{Kingma2015}.
After each interaction episode, $1024$ samples are randomly drawn from the replay buffer for training the network parameters of all components, including the forward model.
The hyper-parameters used were either adapted from HAC~\cite{Levy2019_Hierarchical} or fine-tuned with preliminary experiments.

\section{Experiments}\label{sec:analysis}

We compare the performance of our framework in several goal-based environments with continuous state and action spaces.
All environments provide a sparse extrinsic reward when the goal is reached.
To evaluate our approach, we record the learning performance in terms of successful rollouts in relation to training rollouts.
Therefore, we alternate training (with exploration using $\epsilon$-greedy) and testing rollouts (without exploration) and measure the success rate as the average number of successful testing rollouts within a testing batch.

\subsection{Environments}
Our proposed approach is evaluated in the following simulated environments:
\begin{figure}[!ht]
     \centering
     \begin{subfigure}[b]{0.32\linewidth}
         \centering
          \includegraphics[width=\linewidth]{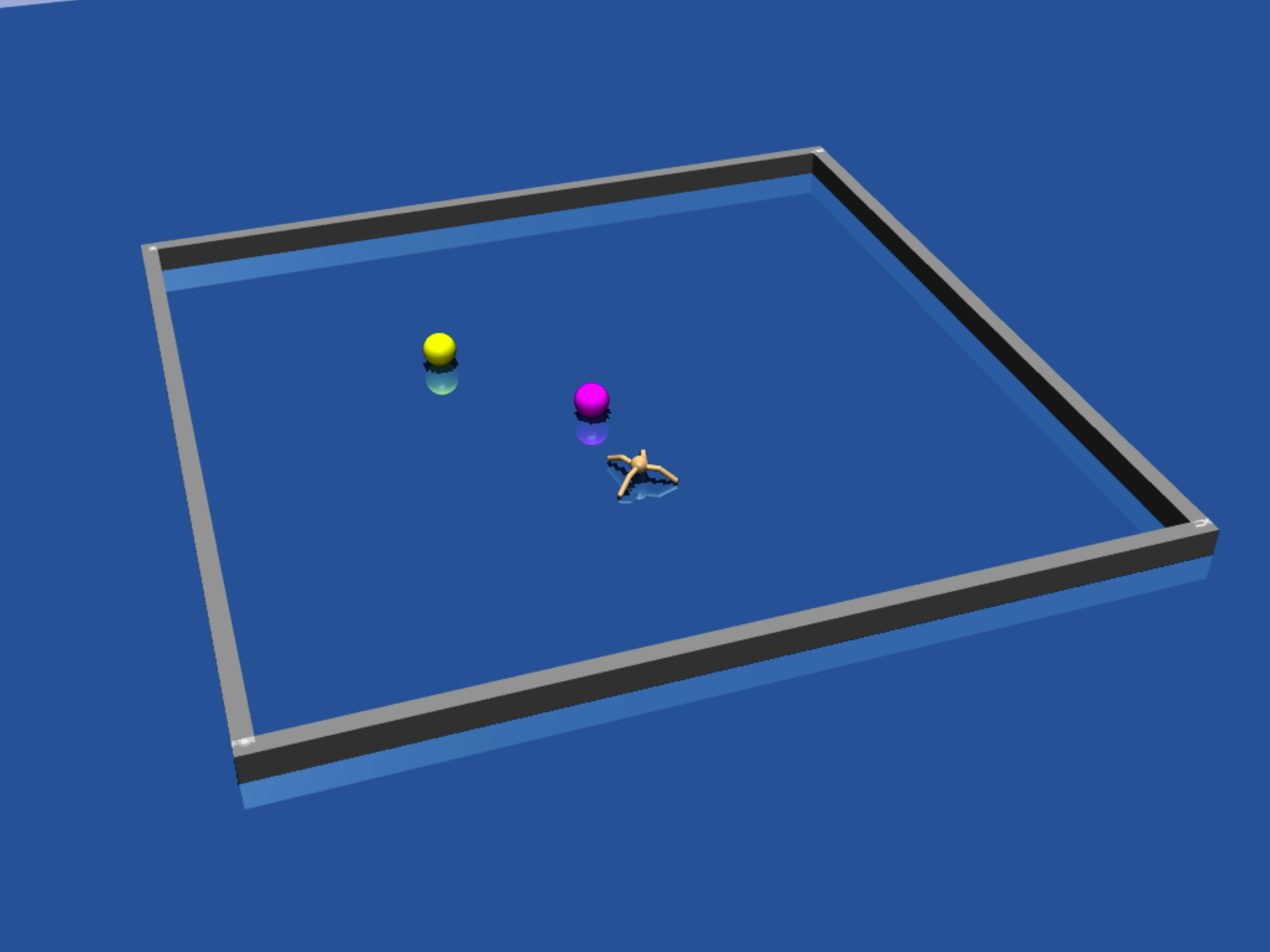}
         \caption{Ant Reacher}
          \label{fig:ant_reach_env}
     \end{subfigure}
     \begin{subfigure}[b]{0.32\linewidth}
         \centering
         \includegraphics[width=\linewidth]{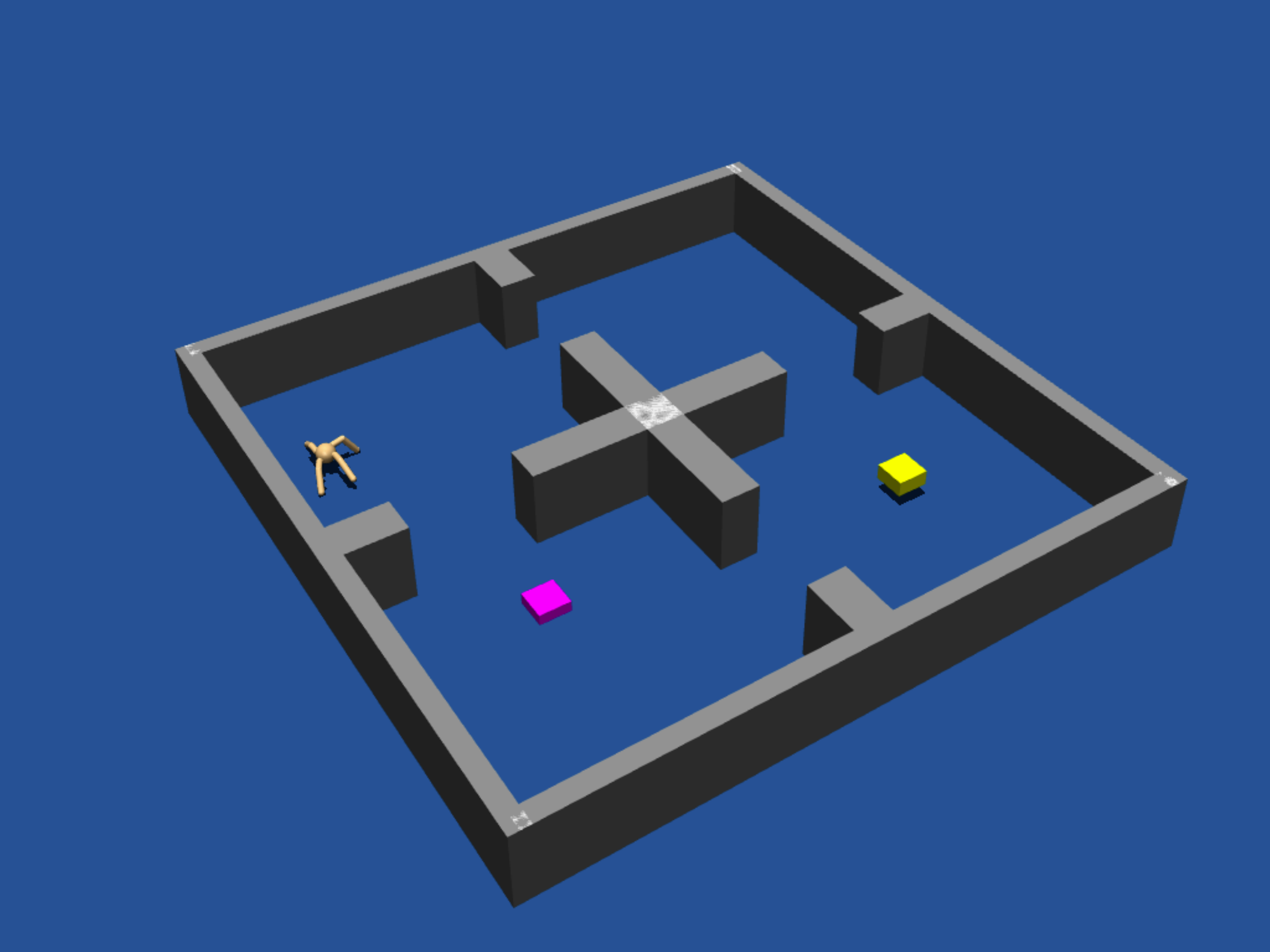}
         \caption{Ant Four Room}
         \label{fig:ant_four_env}
     \end{subfigure}
     \begin{subfigure}[b]{0.32\linewidth}
         \centering
          \includegraphics[width=\linewidth]{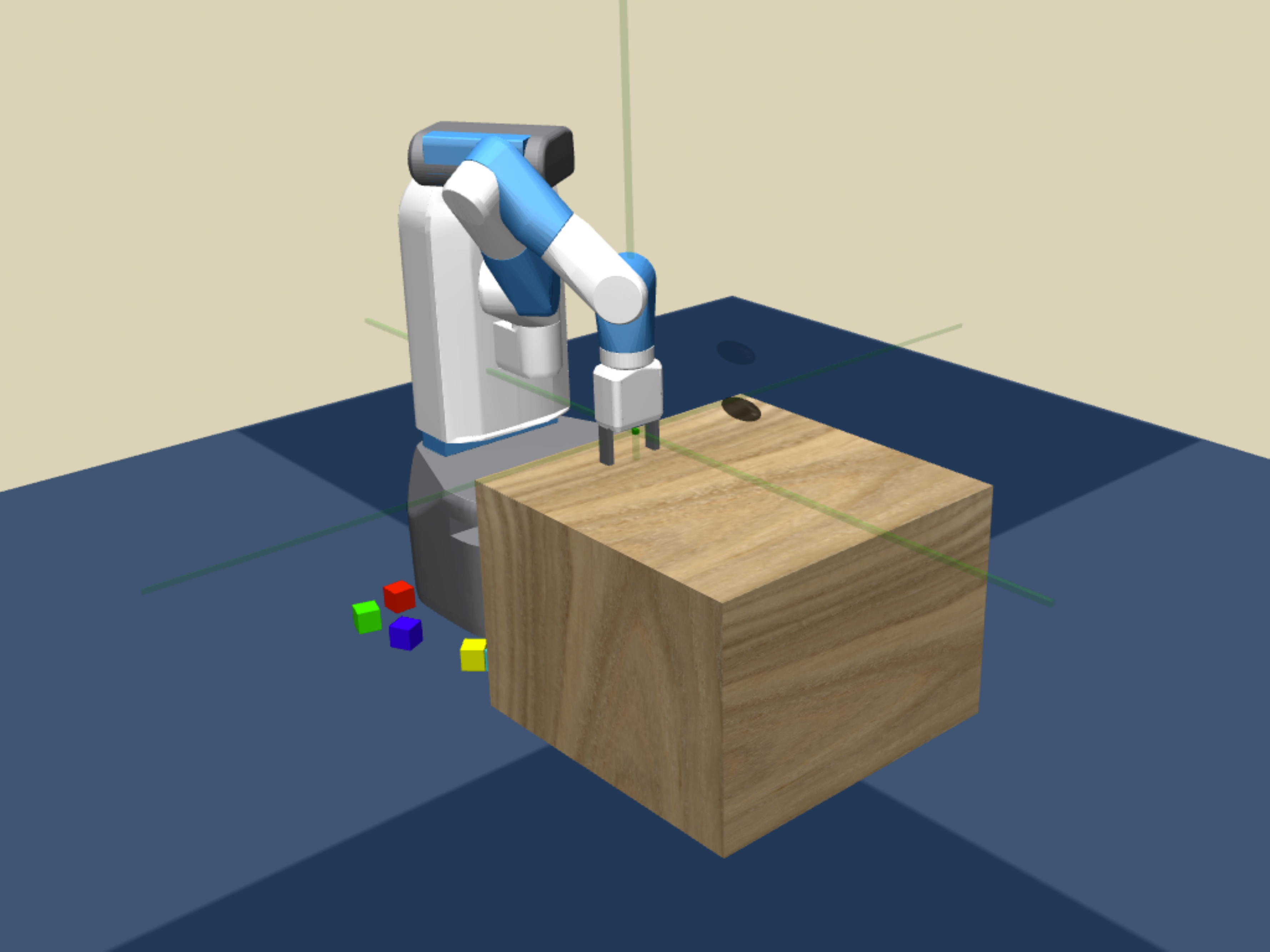}
         \caption{Fetch Reacher}
          \label{fig:reach_fetch_env}
     \end{subfigure}
     \begin{subfigure}[b]{0.32\linewidth}
         \centering
          \includegraphics[width=\linewidth]{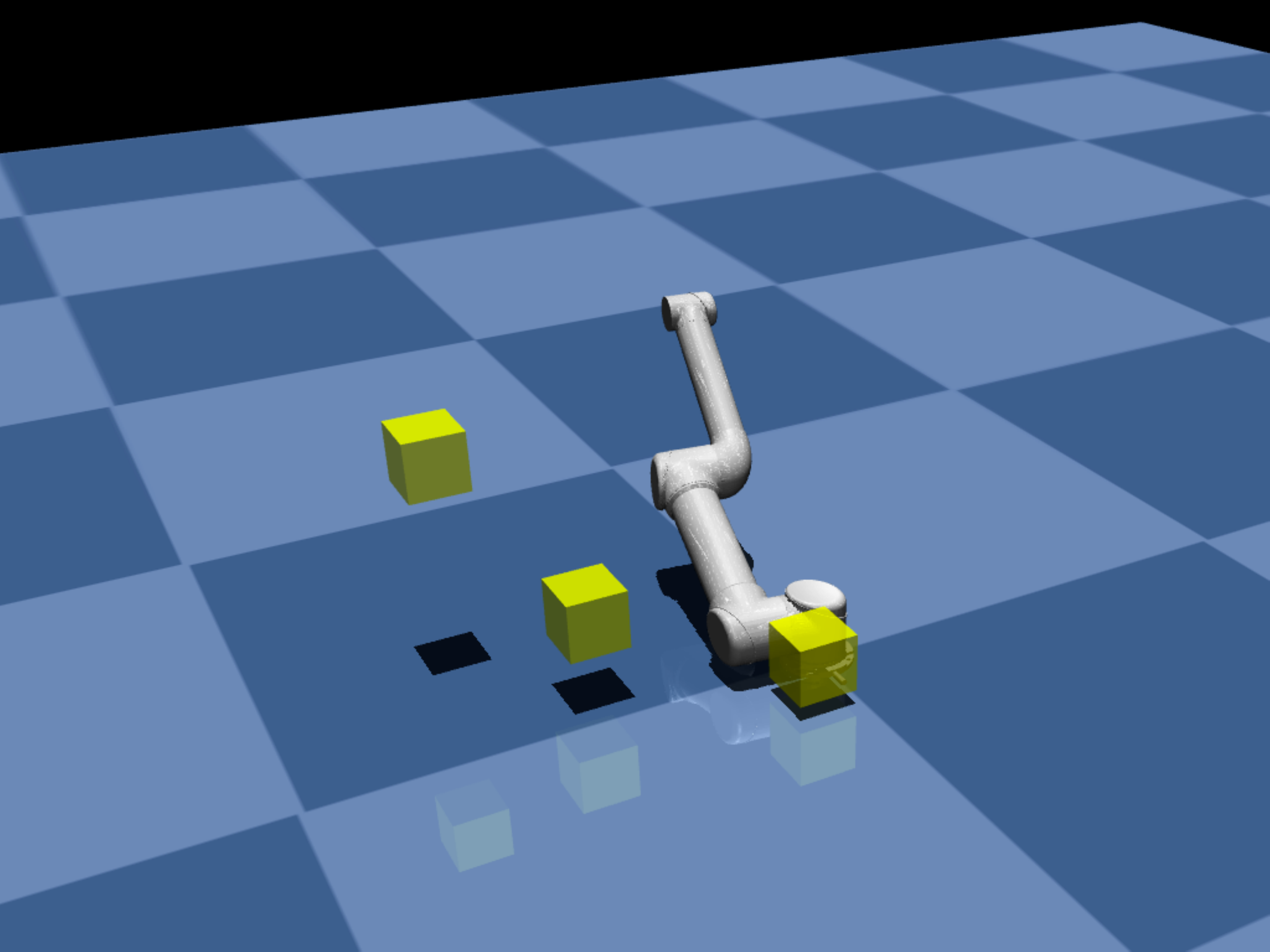}
         \caption{UR5 Reacher}
          \label{fig:UR5}
     \end{subfigure}
     \begin{subfigure}[b]{0.32\linewidth}
         \centering
          \includegraphics[width=\linewidth]{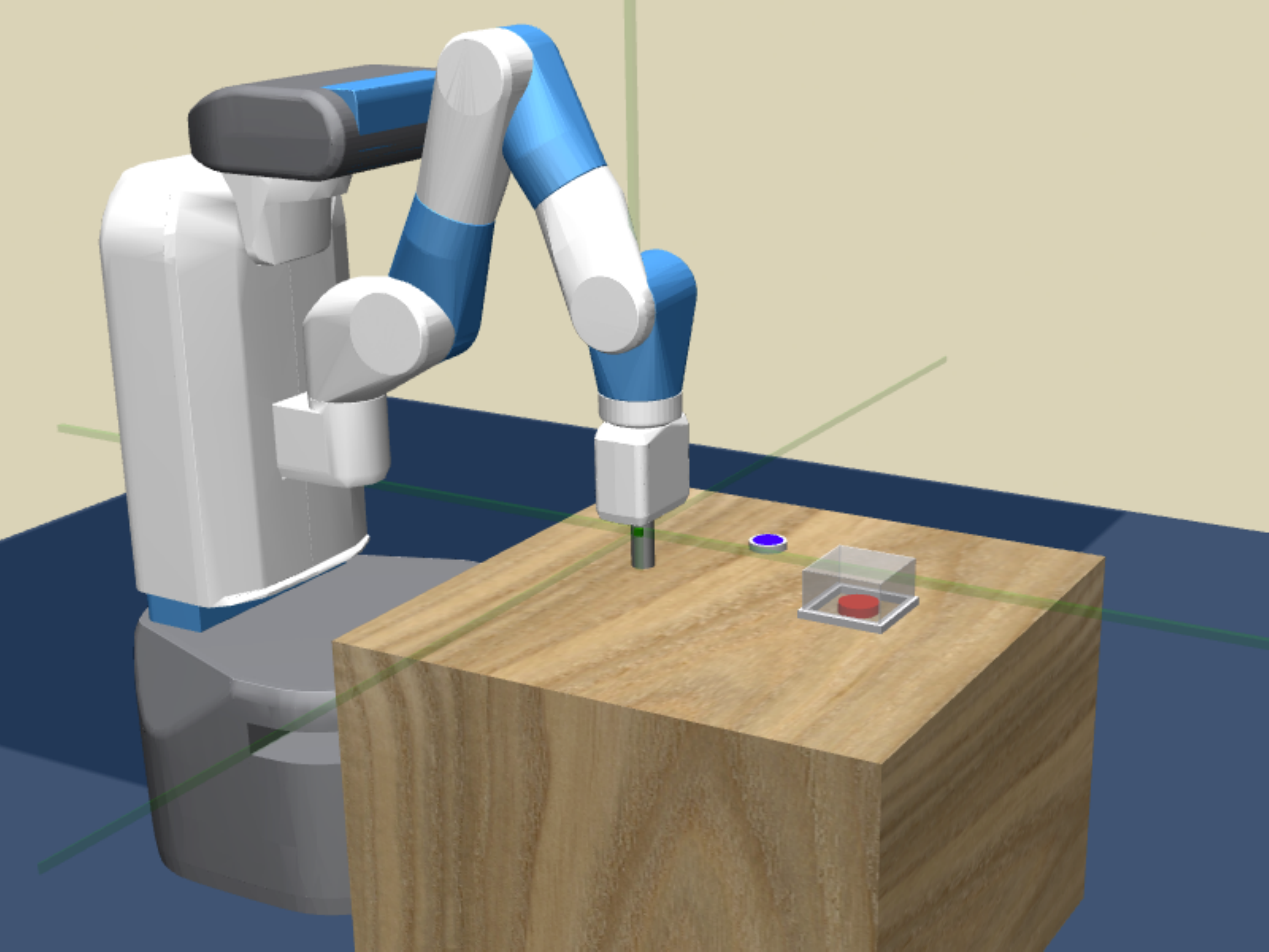}
         \caption{Causal Dependency}
          \label{fig:causal_dep}
     \end{subfigure}
     \begin{subfigure}[b]{0.32\linewidth}
         \centering
          \includegraphics[width=\linewidth]{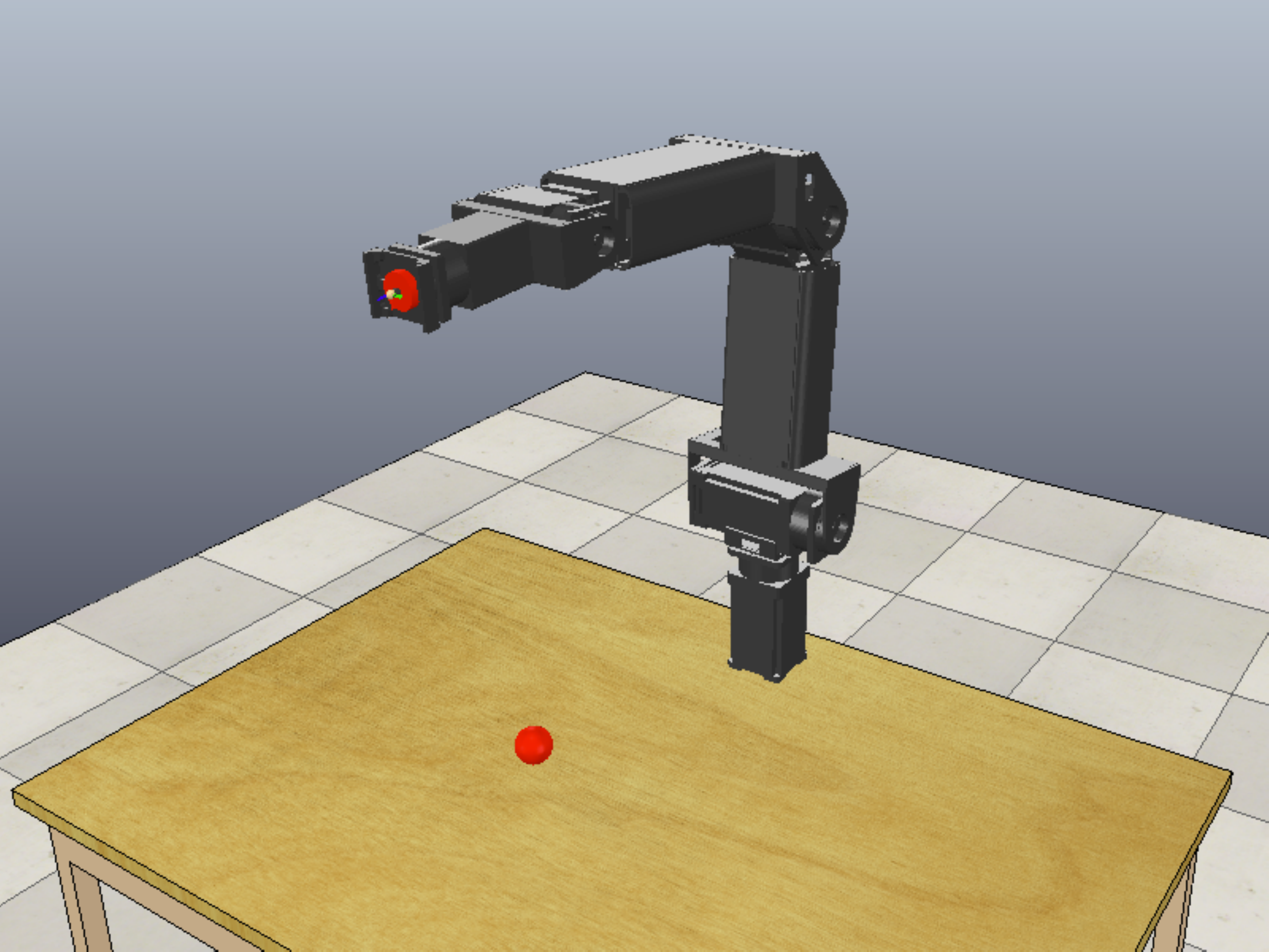}
         \caption{CoppeliaSim Reacher}
          \label{fig:cop_reach}
     \end{subfigure}
     \caption{Simulated environments for experiments}
     \label{fig:envs}
\end{figure}

\begin{itemize}
    \item \textit{Ant reacher:} The \textit{Ant reacher} environment (see \Cref{fig:ant_reach_env}) consists of a four-legged robotic agent that must learn to walk to reach a target location. The action space is based on the joint angles of the limbs, and the observation space consists of the Cartesian locations and velocities of the body parts of the agent. The target location is random Cartesian coordinates of the agent's torso.
    The yellow and pink spheres in the figure indicate the end-goal and subgoal respectively.
    \item \textit{Ant four rooms:} This environment is the same as \emph{Ant reacher}, except that there are walls in the environments that the agent cannot pass (see \Cref{fig:ant_four_env}). The walls form four rooms that are connected by passages to transition from one room to another, increasing the difficulty compared to \emph{Ant reacher}.
    \item \textit{Fetch robot reacher:}
    This reacher environment (see
    \Cref{fig:reach_fetch_env}) is based on an inverse kinematics model that provides a 3D continuous action space. The task of the robot is to move the gripper to a target position (indicated in the figure by the black sphere), defined in terms of Cartesian coordinates.
    \item \textit{UR5 reacher:} This environment consists of the first three DoFs (two shoulder joints and one elbow joint) of a UR5 robotic arm that must reach (feasible) random joint configurations indicated by yellow boxes in \Cref{fig:UR5}. The action space is determined by the angles of the joints, and the state space consists of joint velocities angles.
    \item \textit{Causal Dependency:} The robotic arm of this environment needs to address a simple causal dependency. This dependency is implemented by a button (blue button) that needs to be pressed before a target position (red button) can be reached (cf. \Cref{fig:causal_dep}). The button press opens the lid over the target location so that the arm must first move towards the button and then towards the target location.
    \item \textit{CoppeliaSim Reacher:} This environment is based upon the robot simulation
    CoppeliaSim~\cite{coppeliaSim_2013} and is structured similarly to \textit{Fetch robot reacher}, containing the same task. The task differs from the \emph{Fetch robot reacher} in terms of its goal and observational space. It also makes use of inverse kinematics to reach a target location (red object) seen in \Cref{fig:cop_reach}.
\end{itemize}

\subsection{Results}
\label{sec:results}
\begin{figure}[!ht]
     \begin{subfigure}[b]{0.49\linewidth}
         \centering
         \includegraphics[width=\linewidth]{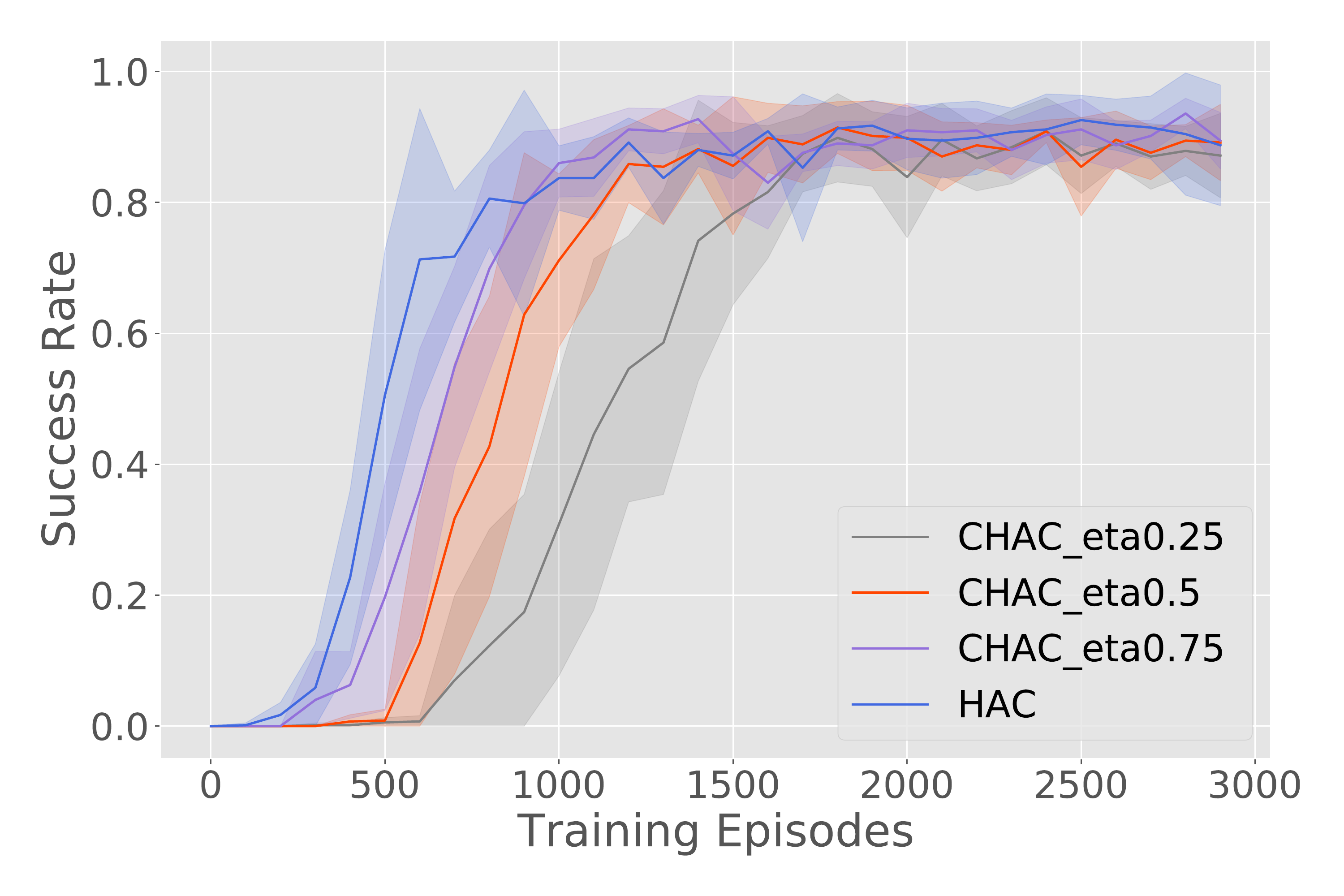}
         \caption{Ant reacher}
         \label{fig:ant_reacher_successrate}
     \end{subfigure}
     \hfill
     \begin{subfigure}[b]{0.49\linewidth}
         \centering
         \includegraphics[width=\linewidth]{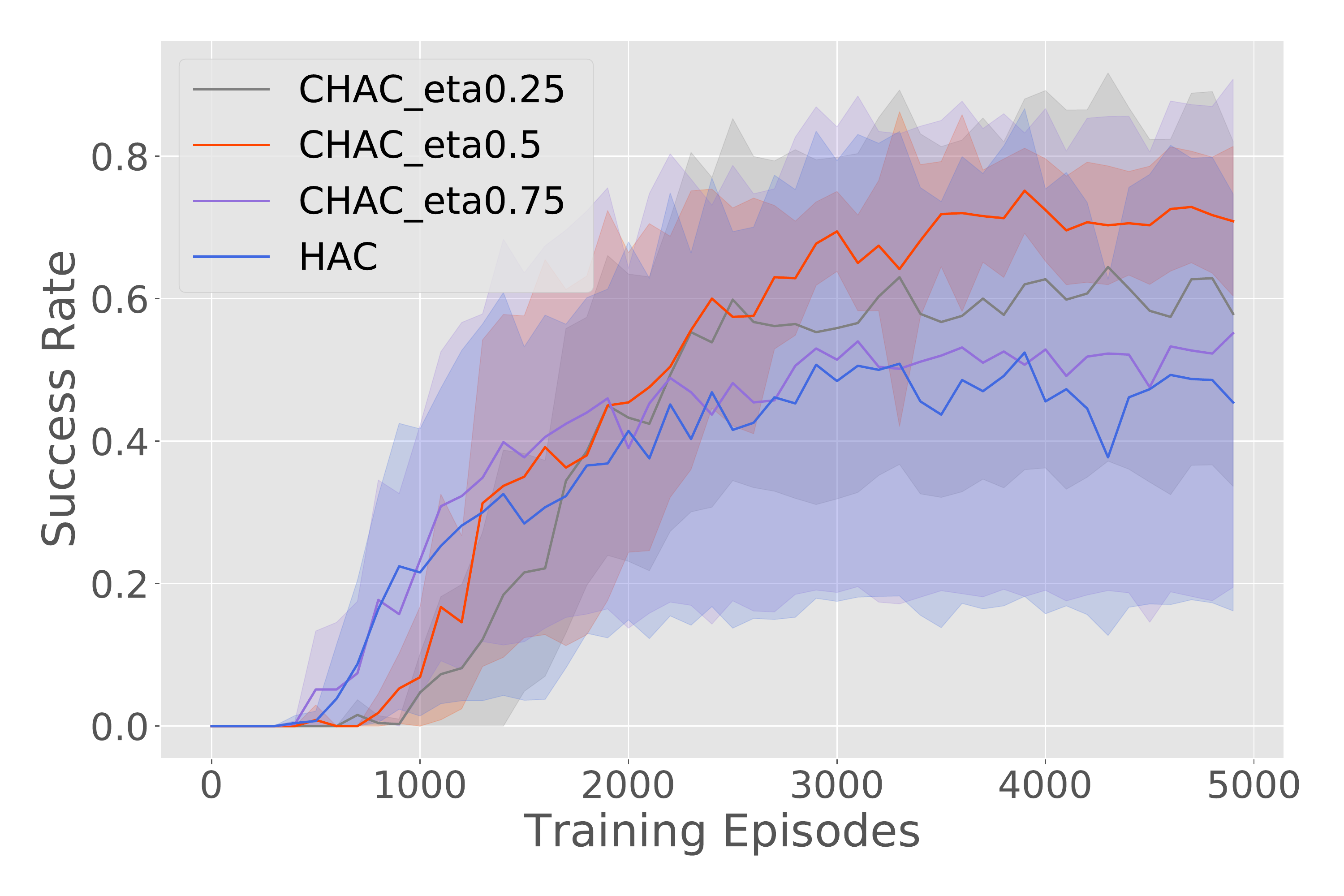}
         \caption{Ant four rooms}
         \label{fig:ant_four_successrate}
     \end{subfigure}

     \begin{subfigure}[b]{0.49\linewidth}
         \centering
         \includegraphics[width=\linewidth]{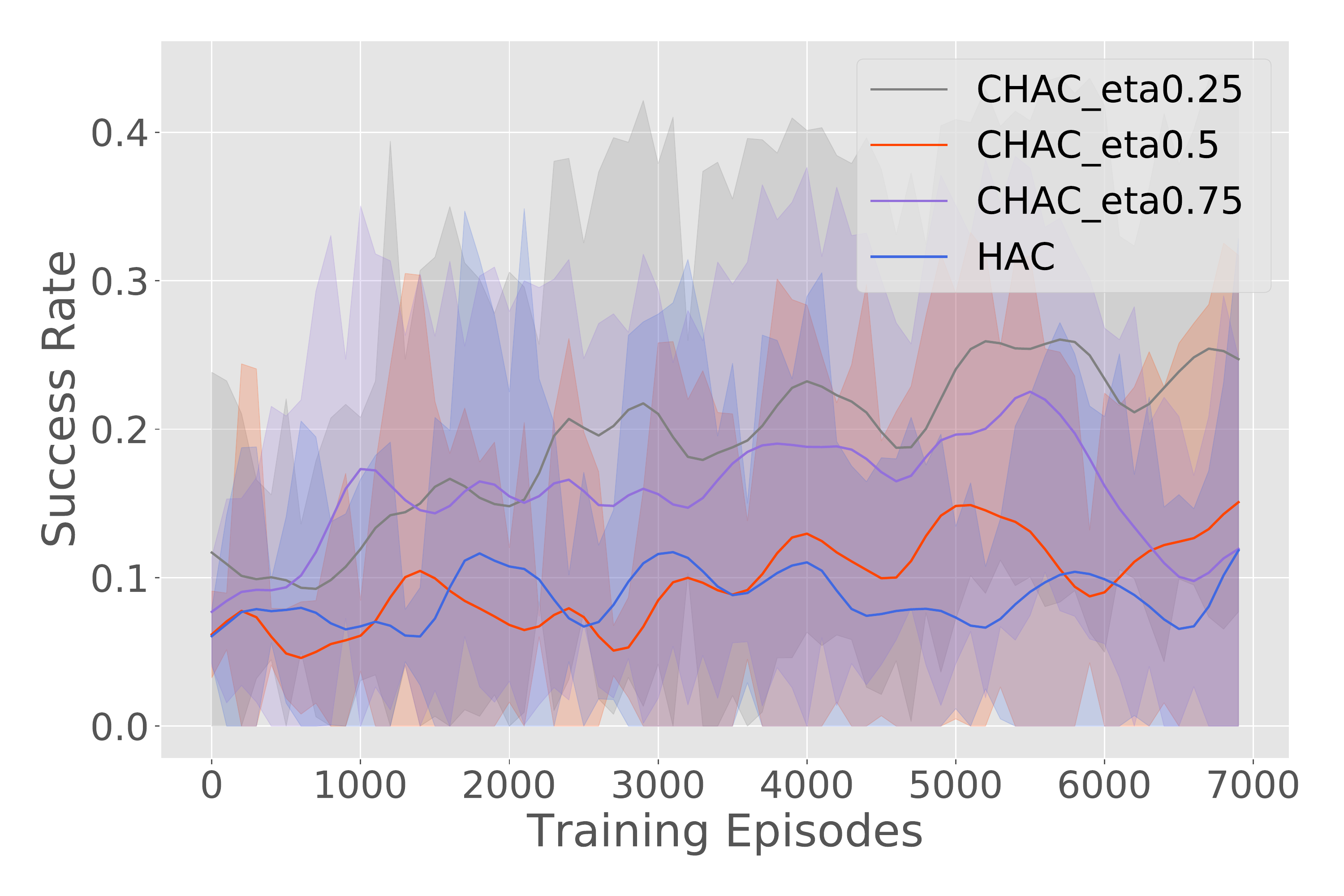}
         \caption{Fetch Reacher}
         \label{fig:block_reacher_successrate}
     \end{subfigure}
     \hfill
     \begin{subfigure}[b]{0.49\linewidth}
         \centering
          \includegraphics[width=\linewidth]{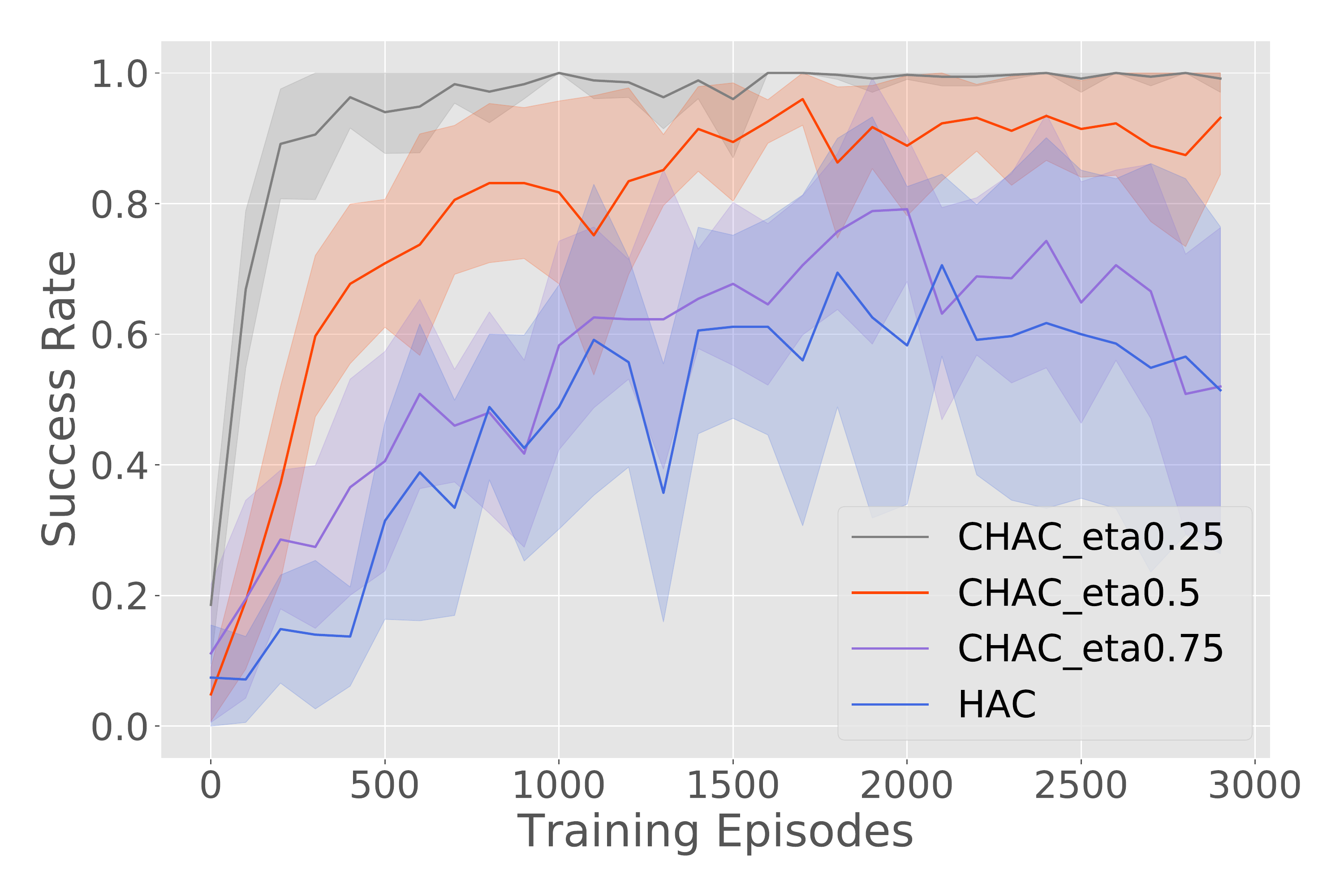}
          \caption{UR5}
          \label{fig:ur5_reacher_successrate}
      \end{subfigure}

      \begin{subfigure}[b]{0.49\linewidth}
         \centering
         \includegraphics[width=\linewidth]{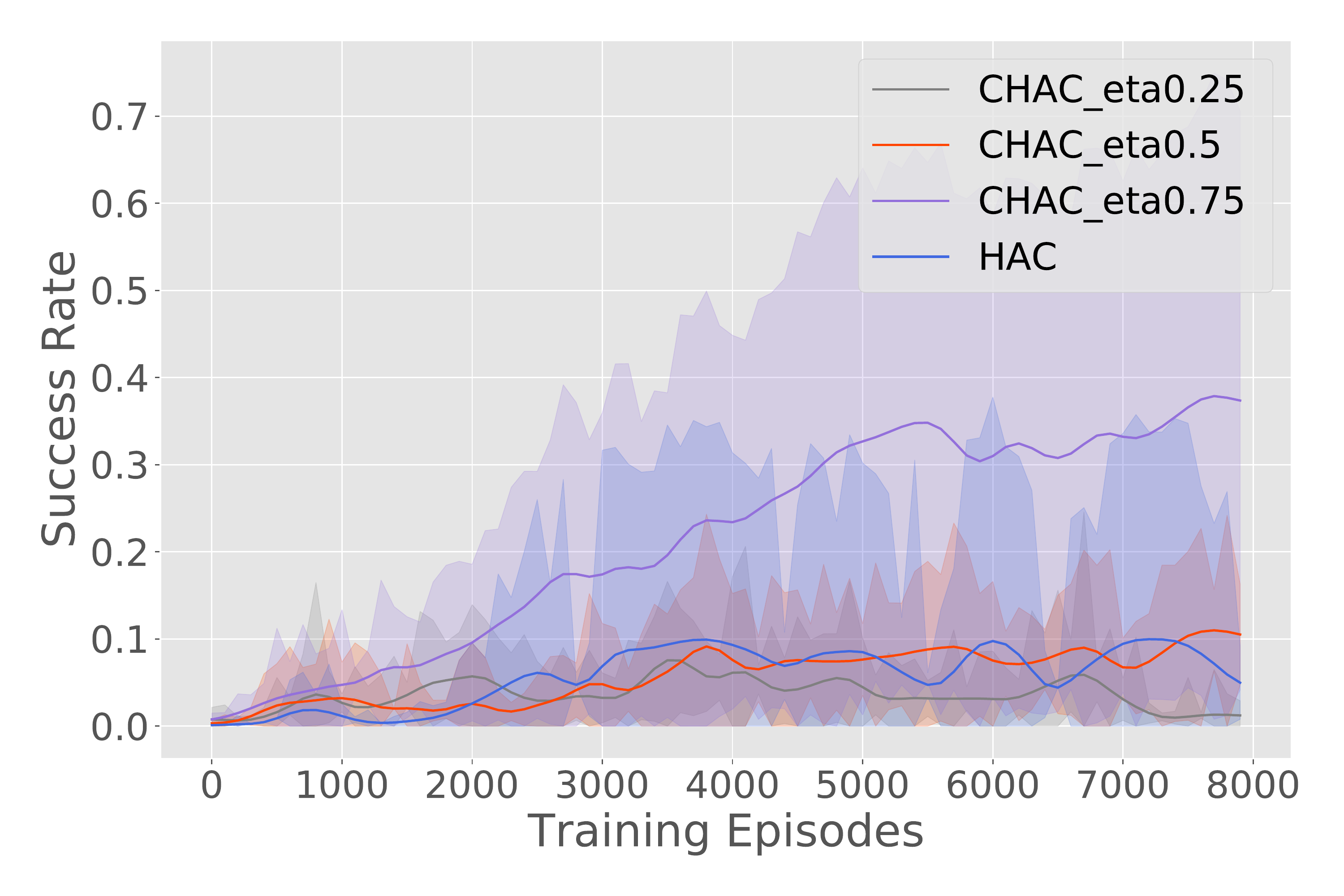}
         \caption{Causal Dependency}
         \label{fig:causal_dep_successrate}
     \end{subfigure}
     \hfill
     \begin{subfigure}[b]{0.49\linewidth}
         \centering
          \includegraphics[width=\linewidth]{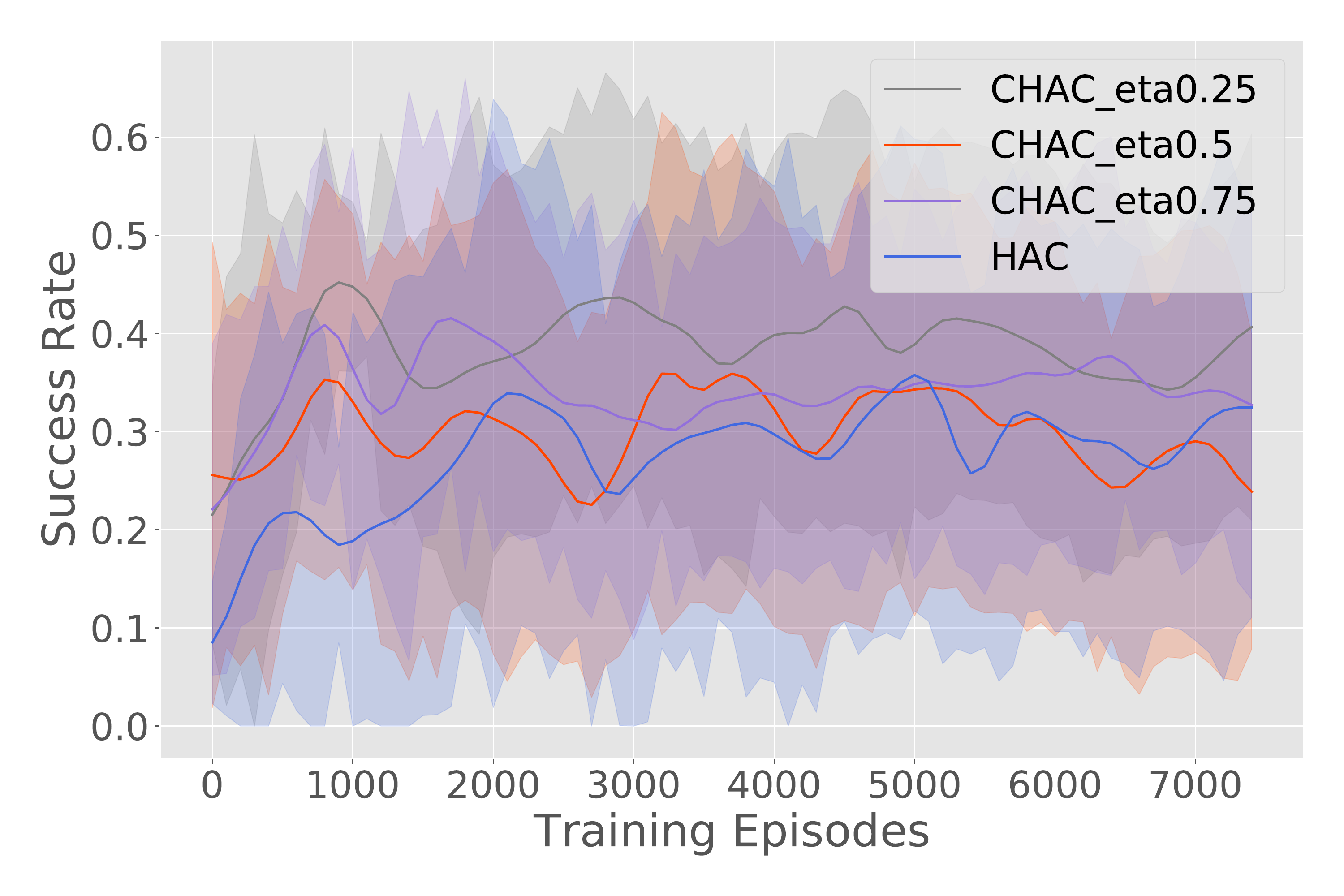}
          \caption{CoppeliaSim Reacher}
          \label{fig:cop_reach_successrate}
      \end{subfigure}

    \caption{Learning performance of the four environments}
    \label{fig:learning_performance}
\end{figure}

Results from \Cref{fig:learning_performance} reveal significant performance gains in terms of the learning progress for most of the investigated environments. For each environment, we use at least seven experiments to calculate the mean. For the shaded area, we use the standard deviation and sometimes apply a bit of smoothing.
The benefit of curiosity differs depending on the task. Hence, we show up four values of $\eta$ for each environment.
For the ant environments (\Cref{fig:ant_reacher_successrate} and \Cref{fig:ant_four_successrate}), curiosity shows different effects. One assumption is that \textit{Ant reacher} is an easier environment and curiosity-driven exploration is not as useful as it is in the more difficult \textit{Ant four rooms}.
For \textit{Ant reacher}, the performance of HAC is quite similar to what CHAC is able to achieve. Both settle in at a mean success rate of 0.9 (cf. \Cref{fig:ant_reacher_successrate}). In \textit{Ant four rooms}, the mean success rate of HAC is between 0.4 and 0.5. When using CHAC with curiosity and $\eta=0.5$, the performance rises and achieves mean success rates between 0.65 and 0.8 (cf. \Cref{fig:ant_four_successrate}).
Within the \textit{Fetch reacher environment}, HAC cannot achieve success rates greater than 0.12. Using CHAC with $\eta \in \{0.25, 0.75\}$ improves the success rates roughly by a factor of 2 (cf. \Cref{fig:block_reacher_successrate}).
The HAC-based UR5 agent achieves a different performance than reported in the paper of HAC~\cite{Levy2019_Hierarchical}\footnote{Our implementation contains a slightly different initialization and gain RPM values for the robot's joints. Nevertheless, the comparison is given.}. However, CHAC speeds up learning by a factor of up to 1.67 with $\eta \in \{0.5, 0.75\}$ (cf. \Cref{fig:ur5_reacher_successrate}).
A performance gain is also achieved within the \textit{Causal Dependency} environment. While HAC fails to learn a good policy, also CHAC struggles with most of its values of $\eta$. Both of them are not able to exceed a mean success rate of 0.12. Except with $\eta=0.75$, CHAC shows up a mean success rate between 0.3 and 0.4 (cf. \Cref{fig:causal_dep_successrate}), resulting in a performance gain of more than 200\%.
The \textit{CoppeliaSim Reacher} shows performance differences right from the start. Even if the training fluctuates, CHAC achieves an improvement roughly 1.5 times better than HAC with $\eta=0.25$.

\section{Conclusion}\label{sec:concl}
Curiosity and the ability to perform problem-solving in a hierarchical manner are two important features of human-level problem-solving and learning.
As a novelty and scientific contribution, this paper presents the first computational approach that combines both features by extending hierarchical actor-critic reinforcement learning with a curiosity-enabled reward function. The level of curiosity is modeled by the prediction error of learnable forward models included in all hierarchical layers.
Our experimental results provide significant evidence that curiosity improves hierarchical problem-solving. Specifically, using the success rate as evaluation metrics, we show that curiosity can more than double the learning performance for the proposed hierarchical architecture and benchmark problems.

\section*{Acknowledgements}
Manfred Eppe, Phuong Nguyen, and Stefan Wermter acknowledge funding by the German Research Foundation (DFG) under the IDEAS project and the LeCAREbot project. We thank Andrew Levy for the productive communication and the publication of the original HAC code.

\def\url#1{} 
\bibliographystyle{splncsnat}
\bibliography{references}

\begin{thebibliography}{31}
\providecommand{\natexlab}[1]{#1}
\providecommand{\url}[1]{\texttt{#1}}
\providecommand{\urlprefix}{}

\bibitem[{Alet et~al.(2020)Alet, Schneider, Lozano-Perez, and
  Kaelbling}]{Alet2020Meta-learningAlgorithms}
Alet, F., Schneider, M.F., Lozano-Perez, T., Kaelbling, L.P.: {Meta-learning
  curiosity algorithms}.
\newblock International Conference on Learning Representations (ICLR) p. online
  (3 2020), \urlprefix\url{https://arxiv.org/abs/2003.05325
  https://openreview.net/forum?id=BygdyxHFDS}

\bibitem[{Andrychowicz et~al.(2017)Andrychowicz, Wolski, Ray, Schneider, Fong,
  Welinder, McGrew, Tobin, Abbeel, and Zaremba~OpenAI}]{Andrychowicz2017}
Andrychowicz, M., Wolski, F., Ray, A., Schneider, J., Fong, R., Welinder, P.,
  McGrew, B., Tobin, J., Abbeel, P., Zaremba~OpenAI, W.: {Hindsight Experience
  Replay}.
\newblock In: Conference on Neural Information Processing Systems (NeurIPS).
  pp. 5048--5058. Curran Associates, Inc. (2017),
  \urlprefix\url{https://papers.nips.cc/paper/7090-hindsight-experience-replay.pdf}

\bibitem[{Bacon et~al.(2017)Bacon, Harb, and Precup}]{Bacon2017_OptionCritic}
Bacon, P.L., Harb, J., Precup, D.: {The Option-Critic Architecture}.
\newblock In: Conference on Artificial Intelligence (AAAI). pp. 1726--1734.
  AAAI Press (2 2017),
  \urlprefix\url{http://aaai.org/ocs/index.php/AAAI/AAAI17/paper/view/14858}

\bibitem[{Botvinick and Weinstein(2014)}]{Botvinick2014}
Botvinick, M., Weinstein, A.: {Model-based hierarchical reinforcement learning
  and human action control}.
\newblock Philosophical Transactions of the Royal Society B: Biological
  Sciences 369(1655) (9 2014),
  \urlprefix\url{http://rstb.royalsocietypublishing.org/cgi/doi/10.1098/rstb.2013.0480}

\bibitem[{Burda et~al.(2019{\natexlab{a}})Burda, Edwards, Pathak, Storkey,
  Darrell, and Efros}]{Burda2019_ICLR}
Burda, Y., Edwards, H., Pathak, D., Storkey, A., Darrell, T., Efros, A.A.:
  {Large-Scale Study of Curiosity-Driven Learning}.
\newblock In: International Conference on Learning Representations (ICLR). p.
  online (2019{\natexlab{a}}),
  \urlprefix\url{https://pathak22.github.io/large-scale-curiosity}

\bibitem[{Burda et~al.(2019{\natexlab{b}})Burda, Edwards, Storkey, and
  Klimov}]{Burda2018}
Burda, Y., Edwards, H., Storkey, A., Klimov, O.: {Exploration by Random Network
  Distillation}.
\newblock International Conference on Learning Representations (ICLR) p. online
  (10 2019{\natexlab{b}}), \urlprefix\url{http://arxiv.org/abs/1810.12894
  https://openreview.net/forum?id=H1lJJnR5Ym}

\bibitem[{Butz(2016)}]{Butz2016}
Butz, M.V.: {Toward a Unified Sub-symbolic Computational Theory of Cognition.}
\newblock Frontiers in psychology 7, 925 (2016),
  \urlprefix\url{http://www.ncbi.nlm.nih.gov/pubmed/27445895
  http://www.pubmedcentral.nih.gov/articlerender.fcgi?artid=PMC4915327
  https://www.frontiersin.org/article/10.3389/fpsyg.2016.00925}

\bibitem[{Colas et~al.(2019)Colas, Fournier, Sigaud, Chetouani, and
  Oudeyer}]{Colas2019}
Colas, C., Fournier, P., Sigaud, O., Chetouani, M., Oudeyer, P.Y.: {CURIOUS:
  Intrinsically Motivated Modular Multi-Goal Reinforcement Learning}.
\newblock In: International Conference on Machine Learning (ICML). pp.
  1331--1340 (2019),
  \urlprefix\url{http://proceedings.mlr.press/v97/colas19a.html}

\bibitem[{Eppe et~al.(2019)Eppe, Nguyen, and Wermter}]{Eppe2019_planning_rl}
Eppe, M., Nguyen, P.D.H., Wermter, S.: {From Semantics to Execution:
  Integrating Action Planning with Reinforcement Learning for Robotic Causal
  Problem-Solving}.
\newblock Frontiers in Robotics and AI 6 (2019),
  \urlprefix\url{https://www.frontiersin.org/articles/10.3389/frobt.2019.00123/full}

\bibitem[{Forestier and Oudeyer(2016)}]{Forestier2016}
Forestier, S., Oudeyer, P.Y.: {Modular active curiosity-driven discovery of
  tool use}.
\newblock In: IEEE International Conference on Intelligent Robots and Systems.
  pp. 3965--3972. IEEE (10 2016),
  \urlprefix\url{http://ieeexplore.ieee.org/document/7759584/}

\bibitem[{Friston et~al.(2011)Friston, Mattout, and Kilner}]{Friston2011}
Friston, K., Mattout, J., Kilner, J.: {Action Understanding and Active
  Inference}.
\newblock Biological Cybernetics 104(1-2), 137--160 (2 2011),
  \urlprefix\url{http://www.ncbi.nlm.nih.gov/pubmed/21327826
  http://www.pubmedcentral.nih.gov/articlerender.fcgi?artid=PMC3491875
  http://link.springer.com/10.1007/s00422-011-0424-z}

\bibitem[{Gottlieb and Oudeyer(2018)}]{Gottlieb2018_curiority_nature}
Gottlieb, J., Oudeyer, P.Y.: {Towards a neuroscience of active sampling and
  curiosity}.
\newblock Nature Reviews Neuroscience 19(12), 758--770 (12 2018)

\bibitem[{Hafez et~al.(2017)Hafez, Weber, and Wermter}]{Hafez2017}
Hafez, M.B., Weber, C., Wermter, S.: {Curiosity-Driven Exploration Enhances
  Motor Skills of Continuous Actor-Critic Learner}.
\newblock In: IEEE International Conference on Development and Learning and
  Epigenetic Robotics (ICDL-EpiRob). pp. 39--46. IEEE (2017),
  \urlprefix\url{http://www.knowledge-technology.info}

\bibitem[{Hester and Stone(2017)}]{Hester2017}
Hester, T., Stone, P.: {Intrinsically motivated model learning for developing
  curious robots}.
\newblock Artificial Intelligence 247, 170--86 (2017),
  \urlprefix\url{http://www.sciencedirect.com/science/article/pii/S0004370215000764}

\bibitem[{Jaderberg et~al.(2017)Jaderberg, Mnih, Czarnecki, Schaul, Leibo,
  Silver, and Kavukcuoglu}]{Jaderberg2017_unreal}
Jaderberg, M., Mnih, V., Czarnecki, W.M., Schaul, T., Leibo, J.Z., Silver, D.,
  Kavukcuoglu, K.: {Reinforcement Learning with Unsupervised Auxiliary Tasks}.
\newblock In: International Conference on Learning Representations (ICLR). p.
  online (11 2017), \urlprefix\url{http://arxiv.org/abs/1611.05397
  https://openreview.net/forum?id=SJ6yPD5xg}

\bibitem[{Jiang et~al.(2019)Jiang, Gu, Murphy, and
  Finn}]{Jiang2019_HRL_Language_Abstraction}
Jiang, Y., Gu, S.S., Murphy, K.P., Finn, C.: {Language as an Abstraction for
  Hierarchical Deep Reinforcement Learning}.
\newblock In: Neural Information Processing Systems (NeurIPS). pp. 9419--9431.
  Curran Associates, Inc. (2019),
  \urlprefix\url{https://sites.google.com/view/hal-demo
  http://papers.nips.cc/paper/9139-language-as-an-abstraction-for-hierarchical-deep-reinforcement-learning.pdf}

\bibitem[{Kingma and Ba(2015)}]{Kingma2015}
Kingma, D.P., Ba, J.L.: {Adam: A Method for Stochastic Optimization}.
\newblock In: International Conference on Learning Representations (ICLR). p.
  online (2015)

\bibitem[{Kulkarni et~al.(2016)Kulkarni, Narasimhan, Saeedi, and
  Tenenbaum}]{Kulkarni2016}
Kulkarni, T.D., Narasimhan, K., Saeedi, A., Tenenbaum, J.B.: {Hierarchical Deep
  Reinforcement Learning: Integrating Temporal Abstraction and Intrinsic
  Motivation}.
\newblock Conference on Neural Information Processing Systems (NeurIPS) pp.
  3675--3683 (2016), \urlprefix\url{http://arxiv.org/abs/1604.06057
  http://papers.nips.cc/paper/6233-hierarchical-deep-reinforcement-learning-integrating-temporal-abstraction-and-intrinsic-motivation.pdf}

\bibitem[{Levy et~al.(2019)Levy, Konidaris, Platt, and
  Saenko}]{Levy2019_Hierarchical}
Levy, A., Konidaris, G., Platt, R., Saenko, K.: {Learning Multi-Level
  Hierarchies with Hindsight}.
\newblock In: International Conference on Learning Representations (ICLR). p.
  online (2019), \urlprefix\url{http://arxiv.org/abs/1712.00948
  https://openreview.net/forum?id=ryzECoAcY7}

\bibitem[{Lillicrap et~al.(2016)Lillicrap, Hunt, Pritzel, Heess, Erez, Tassa,
  Silver, and Wierstra}]{Lillicrap2016_DDPG}
Lillicrap, T.P., Hunt, J.J., Pritzel, A., Heess, N., Erez, T., Tassa, Y.,
  Silver, D., Wierstra, D.: {Continuous Control with Deep Reinforcement
  Learning}.
\newblock In: International Conference on Learning Representations (ICLR). p.
  online (2016), \urlprefix\url{https://arxiv.org/pdf/1509.02971v2.pdf}

\bibitem[{Mnih et~al.(2015)Mnih, Kavukcuoglu, Silver, Rusu, Veness, Bellemare,
  Graves, Riedmiller, Fidjeland, Ostrovski, Petersen, Beattie, Sadik,
  Antonoglou, King, Kumaran, Wierstra, Legg, and Hassabis}]{Mnih2015}
Mnih, V., Kavukcuoglu, K., Silver, D., Rusu, A.A., Veness, J., Bellemare, M.G.,
  Graves, A., Riedmiller, M., Fidjeland, A.K., Ostrovski, G., Petersen, S.,
  Beattie, C., Sadik, A., Antonoglou, I., King, H., Kumaran, D., Wierstra, D.,
  Legg, S., Hassabis, D.: {Human-level control through deep reinforcement
  learning}.
\newblock Nature 518(7540), 529--533 (2 2015),
  \urlprefix\url{http://dx.doi.org/10.1038/nature14236
  http://www.nature.com/articles/nature14236}

\bibitem[{Nachum et~al.(2018)Nachum, Gu, Lee, and Levine}]{Nachum2018_HIRO}
Nachum, O., Gu, S.S., Lee, H., Levine, S.: {Data-Efficient Hierarchical
  Reinforcement Learning}.
\newblock In: Conference on Neural Information Processing Systems (NeurIPS).
  pp. 3303--3313. Curran Associates, Inc. (2018),
  \urlprefix\url{https://sites.google.com/view/efficient-hrl
  http://papers.nips.cc/paper/7591-data-efficient-hierarchical-reinforcement-learning.pdf}

\bibitem[{Pathak et~al.(2017)Pathak, Agrawal, Efros, and
  Darrell}]{Pathak2017_forward_model_intrinsic}
Pathak, D., Agrawal, P., Efros, A.A., Darrell, T.: {Curiosity-driven
  Exploration by Self-supervised Prediction}.
\newblock In: International Conference on Machine Learning (ICML). pp.
  2778--2787. PMLR (2017), \urlprefix\url{https://pathak22.github.io
  http://proceedings.mlr.press/v70/pathak17a.html}

\bibitem[{Pezzulo et~al.(2018)Pezzulo, Rigoli, and
  Friston}]{Pezzulo2018_HierarchicalActiveInference}
Pezzulo, G., Rigoli, F., Friston, K.J.: {Hierarchical Active Inference: A
  Theory of Motivated Control} (4 2018)

\bibitem[{Rohmer et~al.(2013)Rohmer, Singh, and Freese}]{coppeliaSim_2013}
Rohmer, E., Singh, S.P.N., Freese, M.: Coppeliasim (formerly v-rep): a
  versatile and scalable robot simulation framework.
\newblock In: Proc. of The International Conference on Intelligent Robots and
  Systems (IROS) (2013), www.coppeliarobotics.com

\bibitem[{Schaul et~al.(2015)Schaul, Horgan, Gregor, and Silver}]{Schaul2015}
Schaul, T., Horgan, D., Gregor, K., Silver, D.: {Universal Value Function
  Approximators}.
\newblock In: International Conference on Machine Learning (ICML). vol.~37, pp.
  1312--1320. PMLR (2015),
  \urlprefix\url{http://proceedings.mlr.press/v37/schaul15.html}

\bibitem[{Schillaci et~al.(2016)Schillaci, Hafner, and Lara}]{Schillaci2016}
Schillaci, G., Hafner, V.V., Lara, B.: {Exploration Behaviors, Body
  Representations, and Simulation Processes for the Development of Cognition in
  Artificial Agents}.
\newblock Frontiers in Robotics and AI 3, 39 (2016),
  \urlprefix\url{http://journal.frontiersin.org/Article/10.3389/frobt.2016.00039/abstract}

\bibitem[{Schmidhuber(2010)}]{Schmidhuber2010}
Schmidhuber, J.: {Formal Theory of Creativity, Fun, and Intrinsic Motivation
  (1990–2010)}.
\newblock IEEE Transactions on Autonomous Mental Development 2(3), 230--247 (9
  2010), \urlprefix\url{http://ieeexplore.ieee.org/document/5508364/}

\bibitem[{Silver et~al.(2014)Silver, Lever, Hees, Degris, Wierstra, and
  Riedmiller}]{Silver2014DeterministicAlgorithms}
Silver, D., Lever, G., Hees, N., Degris, T., Wierstra, D., Riedmiller, M.:
  {Deterministic Policy Gradient Algorithms}.
\newblock In: International Conference on Machine Learning (ICML). vol.~32, pp.
  387--395 (2014),
  \urlprefix\url{http://proceedings.mlr.press/v32/silver14.html}

\bibitem[{Vezhnevets et~al.(2017)Vezhnevets, Osindero, Schaul, Heess,
  Jaderberg, Silver, and Kavukcuoglu}]{Vezhnevets2017}
Vezhnevets, A.S., Osindero, S., Schaul, T., Heess, N., Jaderberg, M., Silver,
  D., Kavukcuoglu, K.: {FeUdal Networks for Hierarchical Reinforcement
  Learning}.
\newblock In: International Conference on Machine Learning (ICML). vol.~70, pp.
  3540--3549. PMLR (2017), \urlprefix\url{http://arxiv.org/abs/1703.01161
  http://proceedings.mlr.press/v70/vezhnevets17a.html}

\bibitem[{Watters et~al.(2019)Watters, Matthey, Bosnjak, Burgess, and
  Lerchner}]{Watters2019_COBRA}
Watters, N., Matthey, L., Bosnjak, M., Burgess, C.P., Lerchner, A.: {COBRA:
  Data-Efficient Model-Based RL through Unsupervised Object Discovery and
  Curiosity-Driven Exploration} (5 2019),
  \urlprefix\url{http://arxiv.org/abs/1905.09275}

\end{thebibliography}
\end{document}